\documentclass[letterpaper, 10 pt, conference]{ieeeconf}

\usepackage{graphicx}
\usepackage{xcolor}
\definecolor{mydarkblue}{rgb}{0,0.08,0.45}
\usepackage[colorlinks = true,
            linkcolor = mydarkblue,
            urlcolor  = mydarkblue,
            citecolor = mydarkblue,
            anchorcolor = mydarkblue]{hyperref}
\usepackage{amsmath}
\usepackage{amssymb}
\usepackage{mathtools}
\usepackage{bm}
\usepackage{glossaries}
\usepackage{caption}
\usepackage{subcaption}
\usepackage[export]{adjustbox}
\usepackage[linesnumbered,ruled,vlined]{algorithm2e}

\IEEEoverridecommandlockouts
\newacronym{rl}{RL}{Reinforcement Learning}
\newacronym{drl}{DRL}{Deep Reinforcement Learning}
\newacronym{mdp}{MDP}{Markov Decision Process}
\newacronym{pomdp}{POMDP}{Partially Observable Markov Decision Process}
\newacronym{ppo}{PPO}{Proximal Policy Optimization}
\newacronym{sac}{SAC}{Soft Actor-Critic}
\newacronym{crossq}{CrossQ}{CrossQ}
\newacronym{droq}{DroQ}{Dropout Q-Functions}
\newacronym{redq}{REDQ}{Randomized Ensembled Double Q-Learning}
\newacronym{aqe}{AQE}{Aggressive Q-Learning with Ensembles}
\newacronym{utd}{UTD}{Update-To-Data}
\newacronym{cpg}{CPG}{Central Pattern Generator}
\newacronym{jtp}{JTP}{Joint Target Prediction}
\newacronym{bn}{BN}{Batch Normalization}

\let\oldtwocolumn\twocolumn
\renewcommand\twocolumn[1][]{%
  \oldtwocolumn[{#1}{
      \vspace{-1em}
      \begin{center}
        \includegraphics[width=\textwidth]{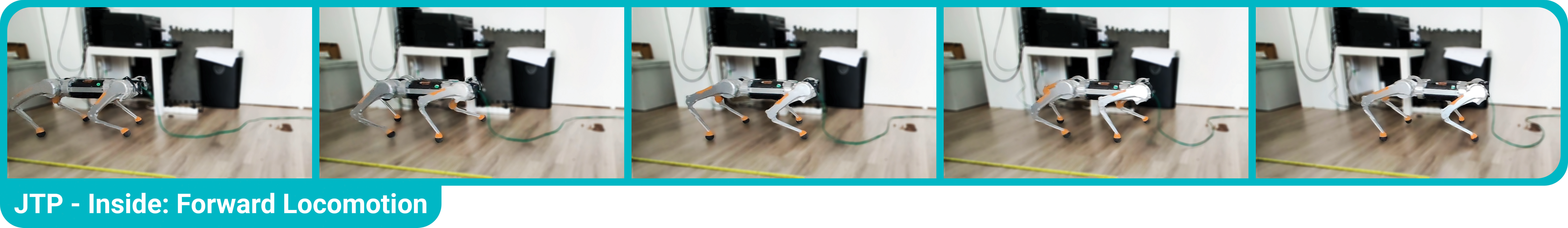}
        
        \vspace{0.5em}
        \includegraphics[width=\textwidth]{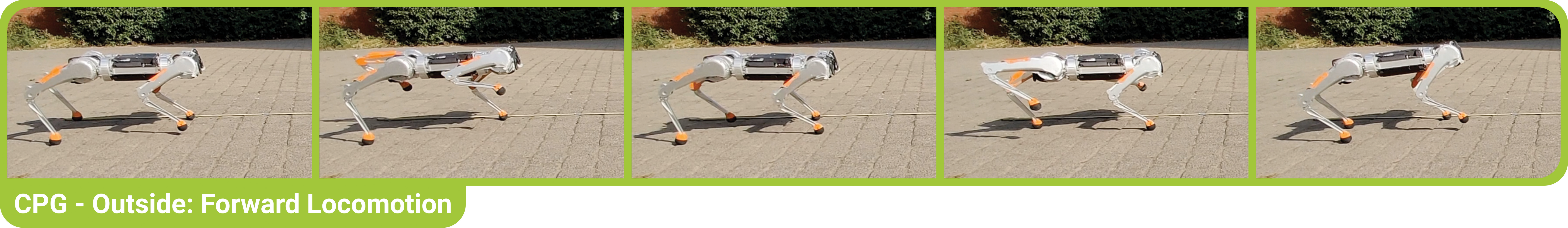}
        
        \vspace{0.5em}
        \includegraphics[width=\textwidth]{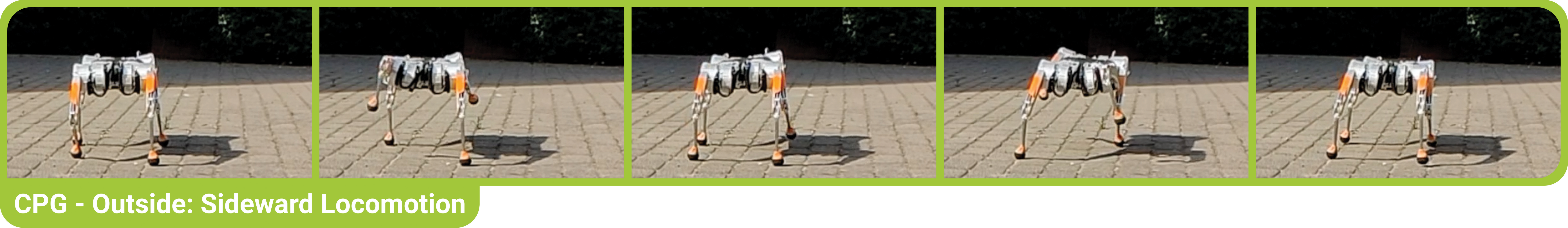}
        \captionof{figure}{
            We demonstrate the effectiveness of the CrossQ algorithm in combination with a Joint Target Prediction (JTP) or Central Pattern Generator (CPG) architecture for learning omnidirectional locomotion on the MAB HoneyBadger quadruped.
        }
        \label{fig:cover_image}
        \vspace{1em}
      \end{center}
  }]
}
\let\ACMmaketitle=\maketitle
\renewcommand{\maketitle}{\begingroup\let\footnote=\thanks \ACMmaketitle\endgroup}

\title{\LARGE \bf
Gait in Eight:
Efficient On-Robot Learning for\\Omnidirectional Quadruped Locomotion
}

\author{
  Nico Bohlinger$^{*,1}$,
  Jonathan Kinzel$^{*,1}$,  
  Daniel Palenicek$^{1,2}$,
  Łukasz Antczak$^{3}$,
  Jan Peters$^{1,2,4,5}$
\thanks{
Funded by the NCN (UMO-2021/43/I/ST6/02711), DFG (PE 2315/17-1), ``Third Wave of AI'' project of the Hessian Excellence Program (HMWK).
\newline $^{*}$Equal contribution.
$^{1}$Department of Computer Science, Technical University of Darmstadt, Germany.
$^{2}$hessian.AI.
$^{3}$MAB Robtics, Poznan, Poland.
$^{4}$German Research Center for AI (DFKI), Research Department: Systems AI for Robot Learning.
$^{5}$Robotics Institute Germany (RIG).
\newline Corresponding author: \tt\small nico.bohlinger@tu-darmstadt.de
}}

\begin{document}

\maketitle
\thispagestyle{empty}
\pagestyle{empty}


\begin{abstract}
On-robot Reinforcement Learning is a promising approach to train embodiment-aware policies for legged robots.
However, the computational constraints of real-time learning on robots pose a significant challenge.
We present a framework for efficiently learning quadruped locomotion in just 8 minutes of raw real-time training utilizing the sample efficiency and minimal computational overhead of the new off-policy algorithm CrossQ.
We investigate two control architectures: Predicting joint target positions for agile, high-speed locomotion and Central Pattern Generators for stable, natural gaits.
While prior work focused on learning simple forward gaits, our framework extends on-robot learning to omnidirectional locomotion.
We demonstrate the robustness of our approach in different indoor and outdoor environments and provide the videos and code for our experiments at: \url{https://nico-bohlinger.github.io/gait_in_eight_website}
\end{abstract}

\section{INTRODUCTION}
Legged robot locomotion has long been an important research area in robotics, as achieving robust, agile, and adaptable gaits in unstructured environments is a challenging yet essential capability for many real-world applications.
Traditional model-based controllers \cite{pratt2006capture, kuindersma2016optimization}, while effective in well-structured settings, often struggle to handle the inherent uncertainties and dynamic variations encountered in real-world terrains.
In contrast, \gls{drl} offers a promising paradigm that allows robots to autonomously acquire locomotion skills directly through interaction with the environment.
In recent years, \gls{drl} has shown remarkable success in learning complex and agile locomotion skills for many different legged robots \cite{bohlinger2024onepolicy, ai2025towards}, such as high-speed running \cite{bellegarda2022robust, margolis2024rapid}, jumping and climbing in parkour-like courses \cite{zhuang2023, cheng2023}, navigating through challenging terrain \cite{agarwal2023legged, zhang2024learning}, and performing handstands and backflips \cite{cheng2023, kim2024stage}.
However, these works rely on scaling up on-policy \gls{drl} algorithms, mainly \gls{ppo} \cite{schulman2017}, through thousands of parallel simulated environments with GPU-based physics engines \cite{rudin2022}.
While a plethora of domain randomization is necessary to zero-shot transfer policies trained in simulation to the real world \cite{tobin2017domain, ji2022concurrent}, this creates a significant embodiment gap between the widely randomized and approximated dynamics of the simulated robot and the specific, nuanced dynamics of the real robot.
On-robot learning promises to bridge the embodiment gap by learning directly on the real system.
This enables the \gls{drl} agent to be aware of its physical embodiment and the specific hardware constraints.
It can continuously adapt to changes, such as wear and tear, battery depletion, hardware modifications, or environmental changes.
Recent advances in off-policy \gls{drl} have made first steps toward this paradigm by improving the learning efficiency enough to enable training quadruped locomotion in real-time directly on the robot \cite{smith2023walk, smith2024grow, levy2024learning}.
However, these works must simplify the learning task to plain forward locomotion with a fixed target velocity.
They only achieve crawling gaits and rely on powerful laptops with dedicated GPUs to run the training process fast enough to be feasible.
Our goal is to lift computational constraints by improving the learning efficiency and update speed with the recently proposed \acrshort{crossq} \gls{drl} algorithm \cite{bhatt2024crossq} and extend the learning task to omnidirectional locomotion, allowing any desired velocity in the $xy$-plane.

Our main contributions are as follows:
\begin{itemize}
    \item We introduce an efficient on-robot learning framework based on \acrshort{crossq}
    \item We learn forward and omnidirectional locomotion while reaching higher maximum velocities, training significantly faster, and doubling the action frequency compared to previous works.
    \item We investigate and compare two control architectures based on \gls{jtp} and \glspl{cpg}, highlighting the trade-offs between achieving aggressive, high-speed gaits and maintaining stable, natural locomotion with high foot clearance.
    \item We provide comprehensive empirical evaluations in simulation and in two real-world environments, demonstrating the practical viability and effectiveness of our framework.
\end{itemize}

\section{PRELIMINARIES}
In this section, we introduce the necessary background and notational foundation for the remainder of the paper.
First, we describe the \gls{rl} framework and efficient algorithms to learn in it.
Then, we introduce the quadruped robot platform used in our experiments in both simulation and the real world.

\subsection{Reinforcement Learning}
We formulate the problem of learning on-robot locomotion as training an \gls{rl} agent that interacts with an environment defined by a \gls{pomdp} $\mathcal{M} = (\mathcal{S}, \mathcal{A}, \mathcal{O}, P, O, R, \gamma)$, where $\mathcal{S}$ is the state space, $\mathcal{A}$ is the action space, $\mathcal{O}$ is the observation space, $P$ is the transition dynamics, $O$ is the observation function, $R$ is the reward function, and $\gamma$ is the discount factor.
Due to partial observability and noisy sensors, the agent does not have access to the true state of the environment $s \in \mathcal{S}$, but instead receives observations $o \in \mathcal{O}$.
To learn a control policy $\pi(a|o)$ that solves the \gls{pomdp}, the agent needs to explore the environment sufficiently.
Therefore, we employ the Maximum Entropy \gls{rl} framework \cite{ziebart2008maximum}.
The goal is to learn a policy that maximizes the expected discounted return while also maximizing the entropy of the policy $J(\pi) = \mathbb{E}_{\tau \sim \pi} \left[ \sum_{t=0}^{\infty} \gamma^t (r_t - \alpha \mathcal{H}(\pi(\cdot|o_t))) \right]$, where $\tau = (o_0, a_0, r_0, o_1, a_1, r_1, \ldots)$ is a trajectory generated by rolling out the policy $\pi$ in the environment and $\mathcal{H}(\pi(\cdot|o_t))$ and $\alpha$ are the entropy of the policy and the temperature parameter, respectively.

\subsection{Efficient model-free off-policy Reinforcement Learning}
\gls{sac} \cite{haarnoja2018soft} is a popular choice of model-free off-policy \gls{drl} algorithms for tasks with continuous state-action spaces.
\gls{sac} is an actor-critic algorithm, formulated for the Maximum Entropy \gls{rl} framework.
As such, it learns a soft Q-function 
\begin{align*}
    \textstyle Q_{\phi}^\pi(o, a) = \mathbb{E}_{\tau \sim \pi} \left[ \sum_{t=0}^{\infty} \gamma^t (r_t - \alpha \log \pi(a_t|o_t)) \right],
\end{align*}
with $o_0 = o$ and $a_0 = a$, 
which models the expected discounted return of a policy $\pi$ when taking action $a$ based on the current observation $o$.
This is done by minimizing the Bellman error $[Q_{\phi}^\pi(o_t, a_t) - r_t - \gamma Q_{\phi}^\pi(o_{t+1}, a_{t+1})]^2$.
Simultaneously, \gls{sac} learns a parameterized policy $\pi_{\theta}(a|o)$, with the objective of maximizing expected discounted return
\begin{align*}
    \textstyle \theta = \arg\underset{\theta}{\max}\; Q_{\phi}^\pi(o, \pi_\theta(o)).
\end{align*}
To reduce the number of environment interactions, authors have mainly proposed to increase the \gls{utd} ratio, which refers to the number of gradient updates performed per agent environment interaction~\cite{nikishin2022primacy, doro2022replaybarrier, chen2021redq, hiraoka2021droq}.
Naturally, this results in increased compute costs and wall-clock time, which can be problematic for on-robot learning in real-time, especially on a constrained compute budget.
Previous work on on-robot learning for quadruped locomotion has relied on the \gls{droq} algorithm~\cite{hiraoka2021droq} with a \gls{utd} ratio of 20 \cite{smith2023walk, smith2024grow} to achieve sample-efficient but compute-intensive learning.

In this work, we build on the recently proposed \acrshort{crossq} algorithm \cite{bhatt2024crossq} that is based on \gls{sac} and achieves state-of-the-art sample efficiency while maintaining the original \gls{utd} of $1$.
The authors achieve this by carefully using \gls{bn}~\cite{ioffe2015batchnorm} within the critic network and removing target networks.
The main insight is to compute the \gls{bn} statistics on the joint current state-action $(o,a)$ and next state-action $(o',a'\sim\pi_\theta(o'))$ distribution.
In practice, this is implemented via a joint forward pass of the current and next state-action batches through the Q-function.

\subsection{Quadruped platform}

\begin{figure}[t]
  \vspace{0.5em}
    \centering
    \includegraphics[width=0.49\linewidth]{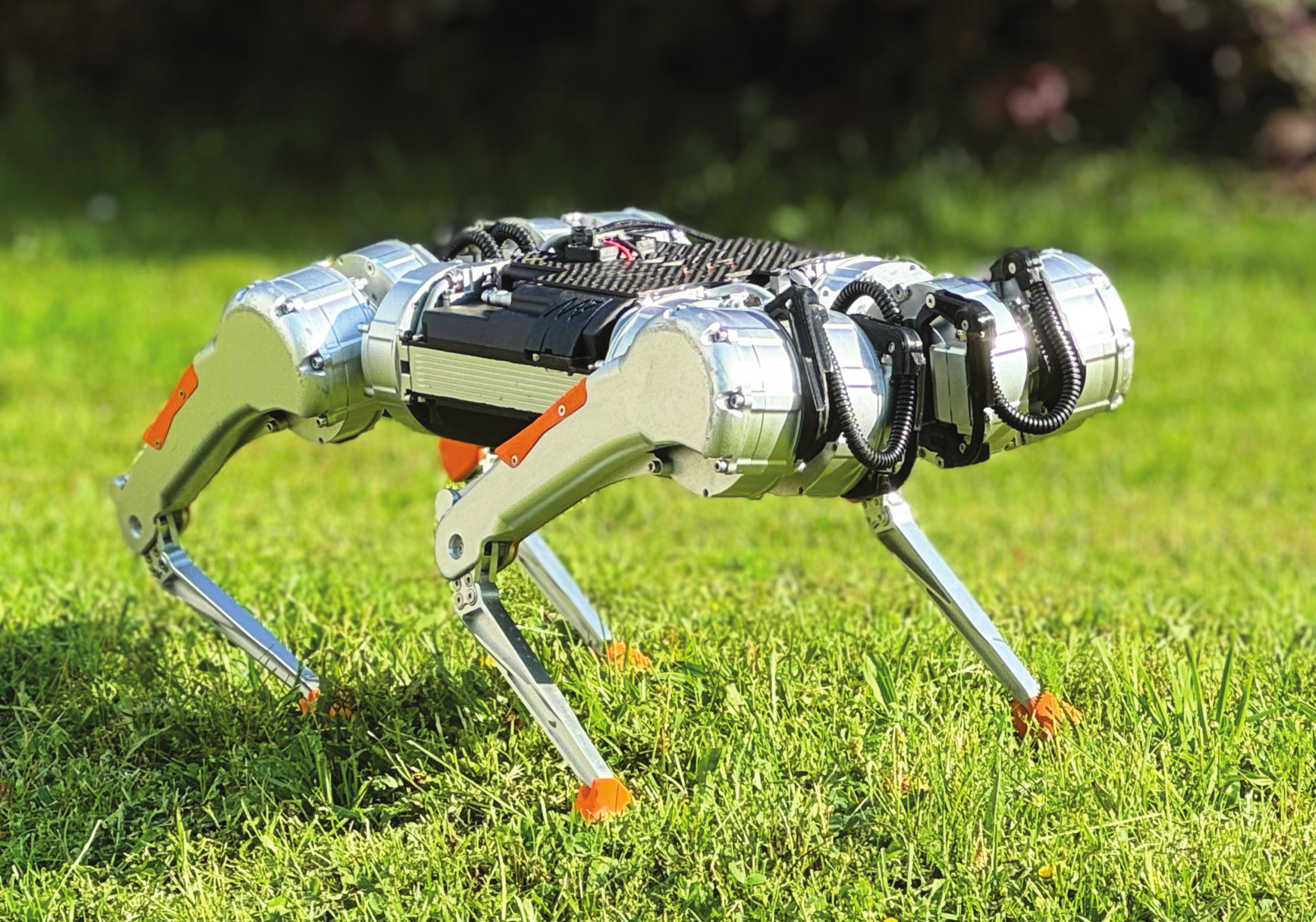}
    \hfill
    \includegraphics[width=0.49\linewidth]{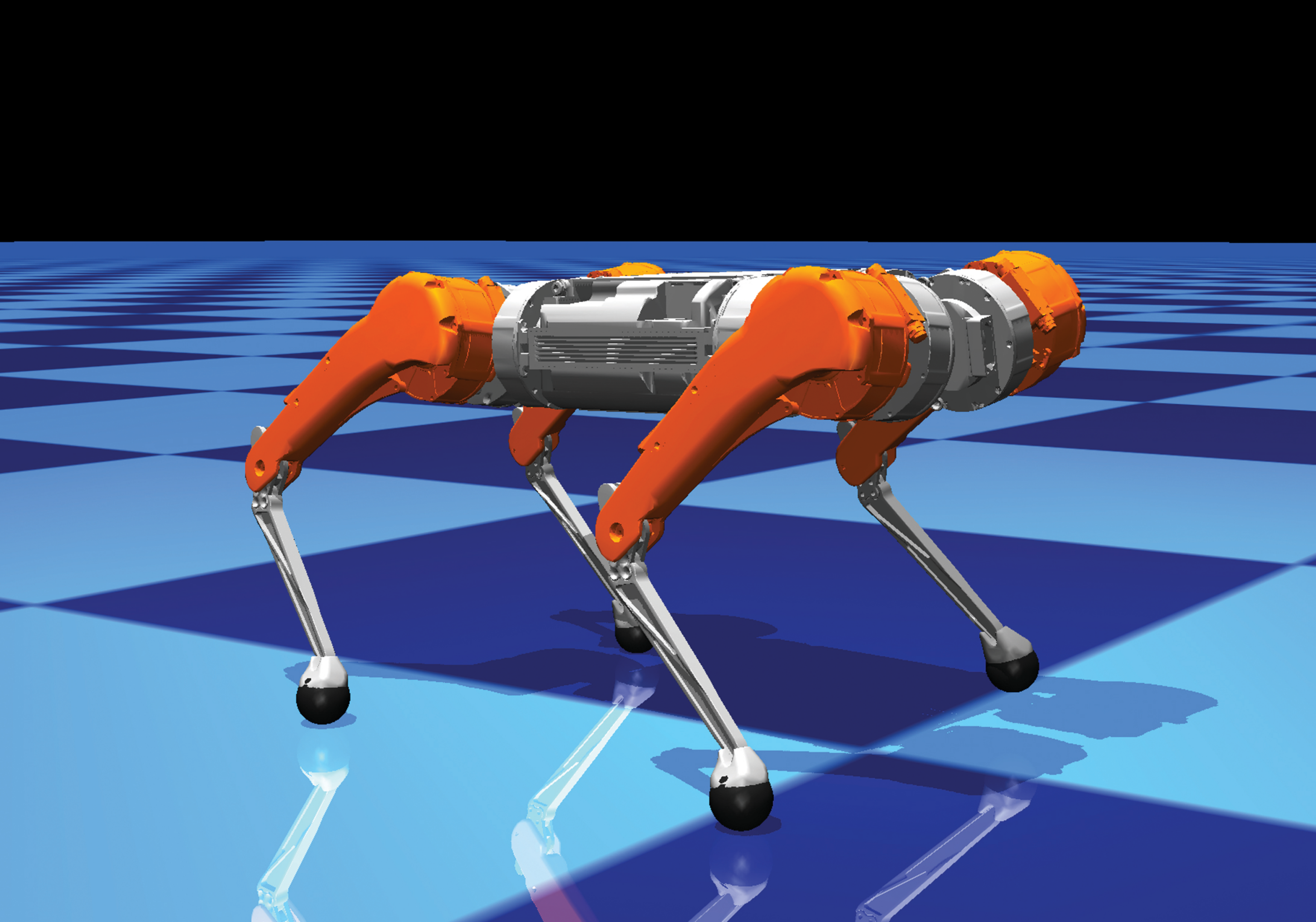}
    \caption{
        MAB Robotics HoneyBadger quadruped robot in the real world (left) and in the MuJoCo simulation (right).
    }
    \label{fig:focused_robot}
    \vspace{-1em}
\end{figure}

All experiments were performed on the HoneyBadger 4.0 quadruped robot from MAB Robotics (Figure~\ref{fig:focused_robot}), a $12 \, \text{DoF}$ platform with three actuated joints per leg.
The robot measures $60 \, \text{cm}$ in length, $40 \, \text{cm}$ in width and height, and has a mass of $12 \, \text{kg}$.
Each joint is driven by a torque-controlled, quasi-direct drive actuator with a 9:1 gear ratio while weighing $0.5 \, \text{kg}$ each.
The actuators are controlled by MAB MD80 servo drives and deliver a nominal torque of $9 \, \text{Nm}$ and a peak of $18 \, \text{Nm}$.
Furthermore, the robot is equipped with a dual-computer system, a VectorNav VN-100 AHRS IMU, and is powered by a $42 \, \text{V}$ Li-Ion battery.
The robot's software is built on ROS 2 that enables the necessary low-level joint control for our experiments.

\section{EFFICIENT ON-ROBOT LEARNING FOR LEGGED LOCOMOTION}
We propose learning legged locomotion directly on the HoneyBadger quadruped robot using the \acrshort{crossq} algorithm and a carefully designed learning framework.
First, we define the task setting and reward design for learning forward and omnidirectional locomotion.
Then, we introduce two control architectures based on \glspl{jtp} and \glspl{cpg} to efficiently learn agile, stable and natural gaits.

\subsection{Locomotion tasks \& reward design}
We first consider the task of learning forward locomotion only as a simplified version of the full locomotion task, since it is commonly used in on-robot locomotion learning~\cite{smith2023walk,smith2024grow,levy2024learning}.
We formulate the task as learning to track a desired $x$-velocity $\bar{v}_x \in \mathbb{R}$ with the robot's trunk.
The reward function is designed to encourage the robot to move in a straight line forward at the target velocity while keeping the body orientation upright and minimizing energy consumption by penalizing high torques.
Ablations on the reward function can be found in appendix \ref{app:ablations}.
Besides tracking a desired velocity, we also consider training the robot to walk forward as fast as possible and change the reward function to simply encourage higher forward velocity while penalizing any velocity in the $y$-direction.
Finally, forward-only locomotion with a fixed target velocity is a common limitation when learning on-robot locomotion, therefore, we extend the task to omnidirectional locomotion by considering random desired velocities in the $xy$-plane $\bar{v}_{xy} \in \mathbb{R}^2$.
Although we omit a target yaw velocity, this setting matches the sim2real literature in robot locomotion more closely \cite{rudin2022learning} and enables a more versatile gait.  
We modify the tracking reward to also consider the $y$-velocity by penalizing any deviation from the target velocity in the $y$-direction.
Table \ref{tab:reward_terms} summarizes the reward functions for the fixed forward, maximum forward, and omnidirectional locomotion tasks, defined by $r_{\text{track-x}}$, $r_{\text{max-x}}$, and $r_{\text{track-xy}}$, respectively.

\begin{figure}[t]
  \vspace{0.5em}
  \centering
  \includegraphics[width=1.0\linewidth]{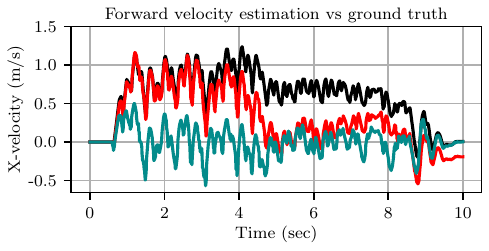}
  \includegraphics[width=1.0\linewidth]{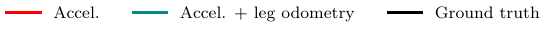}
  \caption{
    Comparison of the linear x-velocity estimation with only integrating the acceleration data from the IMU with a Kalman filter, with the Kalman filter fusing of the accelerations and the leg odometry, and the ground truth.
  }
  \label{fig:velocity_estimation}
  \vspace{-1em}
\end{figure}

\begin{table}[b]
  \vspace{-0.5em}
  \setlength{\tabcolsep}{5pt}
  \def\arraystretch{1.3}
  \caption{
    We build upon the tracking reward $r_{\text{track-x}}$ from \cite{levy2024learning} and add penalty terms for improving the gait quality and energy efficiency.
    $v_x$, $v_y$, and $v_{xy}$ are the x-, y-, and combined xy-velocity of the trunk of the robot, respectively.
    $(R(\theta)^\top\,v_{xy})_y$ is the y component of the trunk velocity rotated into the target direction defined by the target velocity $\bar{v}_{xy}$.
    $\theta$ is the orientation and $\omega$ is the angular velocity of the trunk.
    $\tau$ is the torque applied to the joints.
  }
  \label{tab:reward_terms}
  \centering
  \begin{tabular}{l l}
    \hline
    \textbf{Reward} & \textbf{Term} \\
    \hline
    $r_{\text{track-x}}$ & $\begin{cases}
                1, & \text{for } v_x \in [\bar{v}_x, 2\bar{v}_x]\\
                0, & \text{for } v_x \in (-\infty, -\bar{v}_x] \cup [4\bar{v}_x, \infty) \\
                1 - \frac{|v_x - \bar{v}_x|}{2\bar{v}_x}, & \text{otherwise}
            \end{cases}$ \\
    $r_{\text{max-x}}$ & $v_x - |v_y|$ \\
    $r_{\text{track-xy}}$ & $r_{\text{track-x}} - \Bigl|\Bigl(R(\theta)^\top\,v_{xy}\Bigr)_y\Bigr|$ \\
    $r_{\text{yaw}}$ & $|\omega_{\text{yaw}}|^2$ \\
    $r_{\text{upright}}$ & $|\theta_{\text{pitch, roll}}|^2$ \\
    $r_{\text{energy}}$ & $|\tau|^2$ \\
    \hline
    $r_{\text{total-track-x}}$ & $\max (r_{\text{track-x}} - 0.1 r_{\text{yaw}} - 10 r_{\text{upright}} - 0.0003 r_{\text{energy}}, 0)$ \\
    $r_{\text{total-max-x}}$ & $\max (2 r_{\text{max-x}} - 0.1 r_{\text{yaw}} - 10 r_{\text{upright}} - 0.0003 r_{\text{energy}}, 0)$ \\
    $r_{\text{total-track-xy}}$ & $\max (r_{\text{track-xy}} - 0.1 r_{\text{yaw}} - 10 r_{\text{upright}} - 0.0003 r_{\text{energy}}, 0)$ \\
    \hline
  \end{tabular}
\end{table}

One of the main challenges of training \gls{rl} policies directly on a real robot is the availability of key quantities needed in reward terms and the inherent noise in estimates of these quantities.
Unlike in simulation, where the complete and true state of the robot is readily available, real-world experiments must rely on state estimation techniques that introduce significant uncertainty through sensor noise and may not provide all the necessary information to begin with.
We limit our reward function to rely only on proprioceptive information, such as accurate torque measurements from the joint encoders and noisy data from the onboard IMU.
The orientation of the trunk is estimated by fusing the magnetometer and accelerometer data.
For the linear velocity of the trunk, we use a Kalman filter to fuse the acceleration data with the averaged hip velocity inferred from a leg odometry module that relies on the forward kinematics of the robot.
The linear velocity estimation is the most crucial part of the robot state, as it is the main reward signal driving the learning process, but also the most challenging to estimate accurately due to the lack of foot contact sensors on the robot and the integration of noisy acceleration data.
Figure~\ref{fig:velocity_estimation} shows that integrating only acceleration data works well after an initial calibration but is prone to drift away from the ground truth after a few seconds.
Combining the acceleration data with the leg odometry grounds the estimation.
This prevents significant drift in the estimates, but systematically underestimates the velocity due to the lack of foot contact sensors, which requires the assumption that all four feet are always in contact with the ground.
In general, the learning process is always restricted by the quality of the linear velocity estimation and the ability of the \gls{rl} algorithm to extract useful information from the noisy and biased signal.
A motion capture system could provide a more accurate estimate of the linear velocity \cite{furtado2019comparative} but is only available in a controlled lab setting and, hence, is not suitable for experiments in uncontrolled environments, such as outdoors.

\subsection{Control architecture}
The first control architecture that we consider is the \gls{jtp}, which is commonly used in different variations in legged robot locomotion \cite{rudin2022, smith2023walk, smith2024grow}.
Here, the actions of the policy $a_\text{JTP} \in [-\varphi, \varphi]$ are offset joint angles to a nominal standing position $q^\text{nominal}$ and are clipped to a maximum deviation of $\varphi$.
This ensures a minimal but sufficient joint range for learning a viable gait quickly by reducing the search space of the policy.
The resulting target joint angles $q^\text{target} = q^\text{nominal} + a_\text{JTP}$ are first processed by a filter before being tracked by a PD controller.
The filter manages the Gaussian noise used for policy exploration during the early training phase.
Without filtering, early policies exhibit uncoordinated, jittery movements, leading to inaccurate velocity estimation and oscillations that hurt the learning process and compromise the robot's safety.
Although the low-pass filter is a common choice for smoothing trajectories \cite{lowpassfilter, smith2023walk, smith2024grow}, it also reduces the system's responsiveness to sudden changes, thereby decreasing the robot's potential agility and speed.
Therefore, we employ the One-Euro filter \cite{Casiez2012} as it can significantly enhance responsiveness by balancing the trade-off between low-pass filtering during low velocities and no filtering after a velocity threshold.
Ablations on the choice of the filter can be found in appendix \ref{app:ablations}.

The second control architecture we consider uses the \gls{cpg} framework originating from biology \cite{marder2001central}, where rhythmic patterns are generated by neural circuits in the spinal cord of animals.
In robot locomotion, \glspl{cpg} provide an intuitive formalism to define natural gaits by generating smooth, periodic trajectories for the robot's feet \cite{ijspeert2008central, shao2021learning, kasaei2021}.
We configure a \gls{cpg} to generate a stable in-place trot pattern by defining sinusoidal feet height trajectories $p_{\text{CPG}} = [f(t_i)]_{i=1}^4$ using a spline function
$$
  f(t_i) = \begin{cases}
      h (-2 t_i^3 + 3 t_i^2),    & t_i \in [0, \pi / 2)   \\
      h (2 t_i^3 - 3 t_i^2 + 1), & t_i \in [\pi / 2, \pi)
  \end{cases}
$$
where $t_i$ is the normalized phase of the gait cycle of the $i$-th foot with a fixed frequency.
In case of the trot gait, the phases of the right and left legs are shifted by $\pi / 2$.
The maximum foot height $h$ is set to $0.15 \, \text{m}$, which we empirically validated on the HoneyBadger for robustness on rough terrain with slight inclinations (up to $3^\circ$).
The action of the policy $a_{\text{CPG}} = [x_i, y_i, z_i]_{i=1}^4$ modifies the \gls{cpg} trajectory by predicting an offset position in the Cartesian space for each foot.
Similarly to \gls{jtp}, the predicted offsets are restricted to a Cartesian subspace to reduce the complexity of exploration.
The final feet positions are calculated by $p = p_{\text{CPG}} + a_{\text{CPG}}$, so they can be converted to the corresponding joint angles using analytical inverse kinematics and applied using a PD controller.
While the \gls{cpg} provides a stable in-place trot, the \gls{rl} policy refines this gait, enabling the robot to walk in any direction and dynamically adapt to the environment.

In both control architectures, the agent has access to the following observations: joint angles, joint velocities, previous action, linear and angular accelerations of the trunk, linear and angular velocities of the trunk, desired trunk velocity, and the gravity vector.
As discussed for the reward function, the linear velocities of the trunk are crucial for the learning process, but are the most noisy and biased estimates of the observations.
For the \gls{cpg} approach, the observation space is extended to include the normalized CPG phase variable $[l_1, l_2] \in [0,1]$, which tracks the current progress within the gait cycle for the right and left legs.

\section{EXPERIMENTAL RESULTS}
In this section, we empirically evaluate our framework in both simulation and on the real HoneyBadger robot.
The simulation results compare CrossQ with state-of-the-art off-policy methods in our locomotion setting, focusing on learning efficiency and stability.
Building on these insights, real-world experiments in different environments demonstrate the practical viability of our framework and compare the performance of the proposed control architectures.

\subsection{Simulation}
Before we deploy our learning framework on the real robot, we first evaluate and ablate its performance in simulation.
We use the MuJoCo physics engine \cite{todorov2012mujoco} to simulate the HoneyBadger robot (see Figure~\ref{fig:focused_robot}) and the locomotion tasks on flat terrain with randomized action delays.
We build on the \gls{drl} library RL-X \cite{bohlinger2023rlx} to integrate the simulation environment with the algorithm in JAX, and to run the experiments with 10 seeds for each setting.

\begin{figure}[t]
  \vspace{0.5em}
  \centering
  \includegraphics[width=0.49\linewidth]{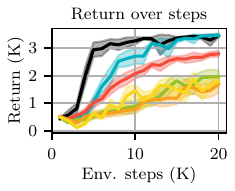}
  \hfill
  \includegraphics[width=0.49\linewidth]{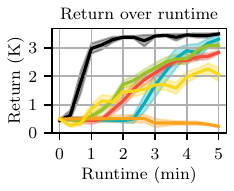}
  \includegraphics[width=0.49\linewidth]{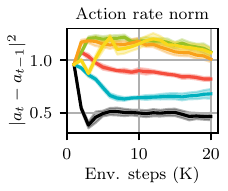}
  \hfill
  \includegraphics[width=0.49\linewidth]{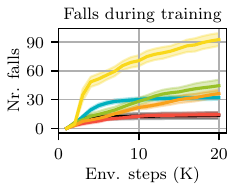}
  \hfill
  \includegraphics[width=\linewidth]{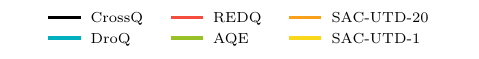}
  \caption{
    Top -- Learning curves of the different algorithms in terms of return over environment steps (left) and runtime (right).
    Bottom -- Number of falls (left) and squared action rate norm (right) during training.
  }
  \label{fig:algorithm_comparison}
  \vspace{-1em}
\end{figure}

First, we perform an ablation study on the learning efficiency of \acrshort{crossq} \cite{bhatt2024crossq} by comparing it to other off-policy algorithms, namely \gls{sac} \cite{haarnoja2018soft}, \gls{aqe} \cite{wu2022aggressive}, \gls{redq} \cite{chen2021redq}, and \gls{droq} \cite{hiraoka2021droq}.
We use the default hyperparameters proposed in the original papers and the same network sizes for all algorithms.
For the \gls{utd} ratio, we use 1 for \acrshort{crossq}, 1 and 20 for \gls{sac}, 5 for \gls{aqe}, and 20 for \gls{redq} and \gls{droq}.
Appendix \ref{app:hyperparameters} summarizes all hyperparameter choices.
We evaluate the algorithms on learning forward locomotion with a target velocity of $\bar{v}_x = 0.5 \, \text{m/s}$ and combine them with the \gls{jtp} control scheme.
Figure \ref{fig:algorithm_comparison} highlights the superior learning speed of \acrshort{crossq} compared to the other algorithms in terms of environment steps and pure training time.
\acrshort{crossq} learns a good locomotion policy after 1 minute of training, while \gls{droq}, the second-best algorithm and used in previous works \cite{smith2023walk, smith2024grow}, requires close to 5 times more training time to reach the same performance.
During the evaluation, the final gait of \acrshort{crossq} appears to be smoother, which is reflected in the squared action rate norm $\sum^{\text{joints}}_j{|a^j_t - a^j_{t-1}|^2}$, which is significantly lower for \acrshort{crossq} compared to the other algorithms.
Importantly, \acrshort{crossq} also achieves the lowest number of falls during training.
The lower action rate norm and the fall rate indicate a better-tempered exploration strategy, which we hypothesize to be due to the removal of target networks, leading to more accurate value estimates.
Fewer falls are highly beneficial for real-world experiments later on, as every fall requires manual intervention and can lead to hardware damage.

\begin{figure}[t]
  \vspace{0.5em}
  \centering
  \includegraphics[width=0.49\linewidth]{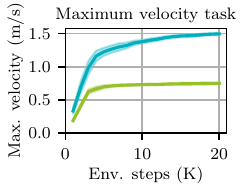}
  \hfill
  \includegraphics[width=0.49\linewidth]{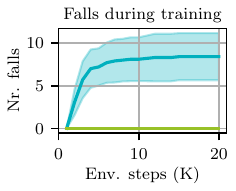}
  \includegraphics[width=0.49\linewidth]{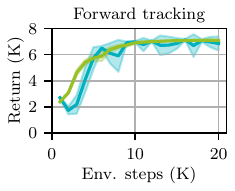}
  \hfill
  \includegraphics[width=0.49\linewidth]{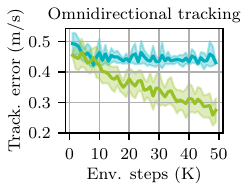}
  \includegraphics[width=\linewidth]{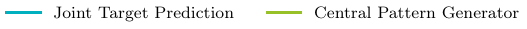}
  \caption{
    Top left -- Average maximum velocity reached in an episode during training on $r_{\text{max-x}}$.
    Top right -- Number of episodes terminated due to falls during the maximum velocity training.
    Bottom left -- Average return on the target forward velocity task.
    Bottom right -- Absolute tracking error of the target velocity on the omnidirectional task.
  }
  \label{fig:jt_vs_cpg_sim_results}
  \vspace{-0.5em}
\end{figure}

Next, we compare the two control architectures: \gls{jtp} and \gls{cpg}.
We train both using \acrshort{crossq} and first evaluate their ability to learn high-speed agile locomotion on the maximum forward velocity task.
Figure \ref{fig:jt_vs_cpg_sim_results} shows that \gls{jtp} reaches a maximum forward velocity of $1.5 \, \text{m/s}$ at the end of training, while the \gls{cpg} achieves only around $0.75 \, \text{m/s}$.
This is expected, as the \gls{jtp} has more direct control over all joints and can learn a more aggressive gait.
However, the learning process with the \gls{cpg} is much more stable and leads to zero falls during training, while the \gls{jtp} approach falls multiple times.
When training the agent on the target forward velocity task with $\bar{v}_x = 0.5 \, \text{m/s}$, both approaches learn to track the target velocity quickly and show a similar end performance.
Using the \gls{cpg} leads to a smaller variance in the learning curves and no falls during training.

Finally, we evaluate the omnidirectional locomotion task with target velocities $\bar{v}_{xy}$ independently sampled from $\mathcal{U}(-0.5, 0.5)$ for $x$ and $y$.
Figure \ref{fig:jt_vs_cpg_sim_results} shows that the \gls{cpg} learns to track target velocities in both directions, while the \gls{jtp} struggles to learn the task at all and falls up to 30 times during training.
It should be noted that we were able to help the \gls{jtp} approach learn by applying a curriculum \cite{bengio2009curriculum} strategy on the sampled target velocities based on the tracking error, but we omit this strategy, as noisy estimates of the tracking error in real-world experiments make it unreliable.
We refer the reader to appendix \ref{app:curriculum} for details.

\begin{figure}[t]
  \centering
  \includegraphics[width=1\linewidth]{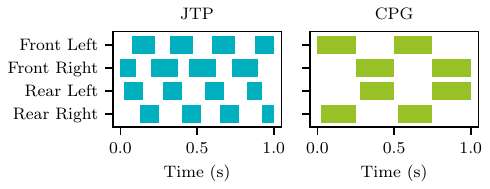}
  \caption{
    Footstep patterns of the learned policies for the \gls{jtp}~(left) and \gls{cpg} (right) approaches.
  }
  \label{fig:footsteps_sim}
  \vspace{-1em}
\end{figure}

In summary, in our experiments the \gls{cpg} approach is more stable and robust. It does not fall at all during training and its initial trot pattern is easily adaptable for the policy to learn omnidirectional locomotion.
The \gls{jtp} approach, on the other hand, learns more agile and aggressive gaits and can achieve higher velocities.
This difference in gait style can also be seen in the different footstep patterns shown in Figure \ref{fig:footsteps_sim}.
The agile \gls{jtp} policies produce more irregular patterns and higher frequency gaits, compared to the \gls{cpg} ones with a fixed gait frequency.

\subsection{Real-world}
We evaluate our learning framework on the real HoneyBadger robot in two settings: a small office environment that provides a smooth, level surface with low friction and no obstacles, and a spacious outdoor environment with uneven cobblestone terrain, high friction, and small obstacles.
The experiments for the forward locomotion task are carried out in both environments, while learning omnidirectional locomotion is only evaluated outdoors due to space constraints in the office environment (see Figure \ref{fig:cover_image}).
The training is performed on a M3 MacBook Pro that is directly connected to the robot via Ethernet and ROS 2. In contrast, previous works rely on a laptop with a dedicated NVIDIA GPU \cite{smith2023walk,smith2024grow}.
Furthermore, we can double the action frequency of our policy to 40 Hz, compared to 20 Hz used in previous works \cite{smith2023walk,smith2024grow}, due to the reduced update time of \acrshort{crossq} with the \gls{utd} ratio of 1 and the removal of the target network.
Like in simulation, we train the policies for 20,000 steps which corresponds to 500 seconds or around 8 minutes of raw training time.
For the omnidirectional locomotion task, we increase the training time to 50,000 steps to handle the increased difficulty of the task.
The actual training duration is significantly longer due to the need for manual intervention, including resetting the robot after falls, reorienting it when it reaches the boundaries of the training area and additional safety triggers that prevent the robot from damaging itself (e.g., joints being close to their angle limits).
We note that using a self-resetting policy that can recover from falls autonomously and is learned prior to the main task could reduce the training time of our experiments \cite{smith2023walk}.
After training, we evaluate the performance of the agent by removing the exploration noise and rolling out the deterministic policy.

\begin{table}[b]
    \vspace{-0.5em}
    \setlength{\tabcolsep}{5pt}
    \def\arraystretch{1.3}
    \caption{
      Comparing real-world experiments with different environments, tasks and control architectures. 
      \textbf{Eval. Vel.}: Maximum velocity reached during evaluation. For omnidirectional locomotion, separate x and y velocities are reported.  
      \textbf{Yaw Ctl.}: Deviation from initial yaw orientation. Very Good ($<$\(10^\circ\)), Good (\(10^\circ - 30^\circ\)), Medium (\(30^\circ - 60^\circ\)), Poor ($>$ \(60^\circ\)).  
      \textbf{Nr. Falls}: Total number of falls during training.  
      \textbf{Durat.}: Time required to complete the training and evaluation.
    }
    \label{tab:real_world_results}
    \centering
    \begin{tabular}{lccccc}
        \hline
        \textbf{App.} & \textbf{Eval. Velocity} & \textbf{Yaw Ctl.} & \textbf{Nr. Falls} & \textbf{Durat.} \\
        \hline
        \multicolumn{5}{l}{\textbf{Office: Forward Locomotion}} \\
        \gls{sac} & $0.013 \, \text{m/s}$   & Poor    & $43$  & $40 \, \text{min}$  \\
        \gls{jtp} & $0.85 \, \text{m/s}$   & Medium    & $15$  & $25 \, \text{min}$  \\
        \gls{cpg} & $0.3 \, \text{m/s}$   & Good      & $38$  & $30 \, \text{min}$  \\
        \hline
        \multicolumn{5}{l}{\textbf{Outside: Forward Locomotion}} \\
        \gls{jtp} & $0.25 \, \text{m/s}$   & Good      & $19$  & $19 \, \text{min}$  \\
        \gls{cpg} & $0.33 \, \text{m/s}$  & Very Good & $39$  & $17 \, \text{min}$  \\
        \hline
        \multicolumn{5}{l}{\textbf{Outside: Omnidirectional Locomotion}} \\
        \gls{jtp} & N/A    & N/A     & $6$  & $25 \, \text{min}$  \\
        \gls{cpg} & x: $0.25 \, \text{m/s}$, y: $0.15 \, \text{m/s}$    & Poor    & $43$ & $33 \, \text{min}$  \\
        \hline
    \end{tabular}
\end{table}

First, we evaluate learning forward locomotion in the indoor office environment.
In addition to combining \acrshort{crossq} with the \gls{jtp} and \gls{cpg} control architectures, we also include a \gls{sac} baseline with a \gls{utd} ratio of 20, a low-pass filter and using the \gls{jtp} approach.
The results of this and all the following real world experiments are summarized in Table \ref{tab:real_world_results}.
The \gls{sac} baseline was only able to reach a very low velocity of $0.013 \, \text{m/s}$ and learned a strategy of heavily jumping with the front legs, leading to harsh movements and poor grip on the ground.
This resulted in the robot being unable to learn a straight and stable gait.
The training took up to 40 minutes to complete, due to the 20 Hz action frequency limited by the high \gls{utd} ratio and a total of 43 falls during training.
\acrshort{crossq} with the \gls{jtp} approach and a One-Euro filter was able to reach a maximum velocity of $0.85 \, \text{m/s}$ after completing the training in 25 minutes with only 15 falls, which is the fastest gait learned in a few minutes directly on a quadruped robot to our knowledge.
The agent initially took small steps while maintaining balance, gradually improving its gait, and increasing its step size over time.
Although the agent developed a fast-paced trotting strategy, it struggled with occasional backward falls and walking in a straight line.
\acrshort{crossq} with the \gls{cpg} reached a maximum velocity of $0.3 \, \text{m/s}$ after 30 minutes of training with 38 falls.
The \gls{cpg} triggered our safety constraints regularly, which led to many unnecessary falls, nevertheless the agent was able to learn a straight and stable gait with high foot clearance.

Next, we test the forward locomotion task in the outdoor environment, which introduces additional complexity due to uneven cobblestones, increased friction, and small obstacles such as curbs.
The \gls{jtp} approach reached a maximum velocity of $0.25 \, \text{m/s}$ after 19 minutes of training with 19 falls.
The agent initially focused on balancing its trunk by deliberately falling backward to prevent tipping forward, but over time leveraged front-leg coordination to maintain stability.
This initial focus on balance led to a slower final gait speed, but the agent was able to adapt to terrain variations effectively.
The \gls{cpg} approach reached a maximum velocity of $0.33 \, \text{m/s}$ after 17 minutes of training with 39 falls.
The final policy showed robustness against environmental disturbances and avoided unnecessary safety activations after an initial phase of struggle with inclinations.

The omnidirectional locomotion experiment introduced random target velocities in both the $x$- and $y$-direction.
Due to the strong noise and drifting in the linear velocity estimation, curriculum learning was not feasible, requiring the agent to adapt without a difficulty progression for the target velocities.
Like in simulation, the \gls{jtp} was unable to learn with the full range of target velocities in the $xy$-plane and failed to achieve any meaningful directional movement, resorting to a standing behavior.
The \gls{cpg} achieved a maximum velocity of $0.25 \, \text{m/s}$ in the $x$-direction and $0.15 \, \text{m/s}$ in the $y$-direction after 33 minutes.
After overcoming early instabilities that resulted in 43 falls, the agent adapted to the outdoor environment, achieving forward, left, and right movements during the evaluation.
However, the learned gait was not perfectly straight and the maximum velocities fell short of the target velocities, leaving room for future work.

In summary, like in simulation, the \gls{jtp} control architecture proved to be more agile, while the gait of \gls{cpg} looked more natural and had better yaw control.
But unlike in simulation, the \gls{cpg} suffered from triggering the joint limit safety constraints, leading to more terminations during training.
Depending on the environment and the desired locomotion task, both control architectures have their advantages and disadvantages, with a trade-off between agility and stability.

\section{CONCLUSION}
In this work, we presented a framework for efficiently learning quadruped locomotion directly on the HoneyBadger quadruped robot using the \acrshort{crossq} algorithm.
Our approach leverages \acrshort{crossq}'s sample efficiency and minimal compute overhead to achieve maximum velocities of up to $0.85 \, \text{m/s}$ in just 8 minutes of raw training.
We combine \acrshort{crossq} with two control architectures: a \gls{jtp} scheme for agile, high-speed gaits and a \gls{cpg} scheme for stable, natural gaits.
Lastly, we extended on-robot locomotion learning to omnidirectional locomotion with different target velocities in the $xy$-plane.
Our real-world experiments in indoor and outdoor environments showed the practicality of our framework and the robustness of the learned policies to terrain variations and sensor noise.
Future work will focus on improving the linear velocity estimation with visual-inertial odometry~\cite{aditya2025robust}, exploring visual observations, algorithmic improvements of the base algorithm~\cite{palenicek2025scaling} and fine-tuning powerful pre-trained policies from simulation to adapt to new environments and the specific robot embodiment that is changing through wear and tear or hardware modifications.




\appendix

\subsection{Learning and filtering hyperparameters}
\label{app:hyperparameters}
We summarize the hyperparameters for the different off-policy algorithms used for the experiments in Table \ref{tab:hyperparameter_configurations}.
The parameters for the action filtering are listed in Table \ref{tab:filter_parameters}.

\begin{table}[h]
  \setlength{\tabcolsep}{5pt}
  \def\arraystretch{1.3}
  \caption{Hyperparameter Configurations}
  \label{tab:hyperparameter_configurations}
  \centering
  \begin{tabular}{lcccccc}
    \hline
    \textbf{Parameter} & \textbf{SAC} & \textbf{SAC-20} & \textbf{DroQ} & \textbf{REDQ} & \textbf{AQE} & \textbf{CrossQ} \\
    \hline
    Learn. Rate & 0.003 & 0.003 & 0.001 & 0.0003 & 0.0003 & 0.005 \\
    Batch Size & 256 & 256 & 256 & 256 & 256 & 128 \\
    Frequency & 20 Hz & 20 Hz & 20 Hz & 20 Hz & 20 Hz & 40 Hz \\
    Neurons & \multicolumn{6}{c}{256, 256} \\
    Nr. critics & 2 & 2 & 2 & 10 & 10 & 2 \\
    Gamma & \multicolumn{6}{c}{0.99} \\
    Optimizer & \multicolumn{6}{c}{Adam} \\
    \gls{utd} ratio & 1 & 20 & 20 & 20 & 5 & 1 \\
    \hline
  \end{tabular}
  \vspace{-0.5em}
\end{table}

\begin{table}[h]
  \setlength{\tabcolsep}{5pt}
  \def\arraystretch{1.3}
  \caption{Filter Parameters}
  \label{tab:filter_parameters}
  \centering
  \begin{tabular}{lccc}
    \hline
    \textbf{Parameter} & \textbf{None} & \textbf{Low-pass filter} & \textbf{One-Euro filter} \\
    \hline
    mincutoff & -- & 0.4 & 2.5 \\
    beta & -- & -- & 0.1 \\
    dcutoff & -- & -- & 100 \\
    \hline
  \end{tabular}
  \vspace{-0.5em}
\end{table}

\subsection{Omnidirectional curriculum}
\label{app:curriculum}
To enable the \gls{jtp} approach to learn omnidirectional locomotion, we introduce a curriculum strategy.
The curriculum is designed to systematically guide the learning agent from easy-to-learn backward locomotion to stable omnidirectional movement in the $xy$-plane.
We model movement directions as a circular space, partitioned into two half-circles, representing leftward and rightward directions.
Each half-circle is further divided into multiple bins, corresponding to incremental directional expansions.
Initially, the agent is trained exclusively in backward movement, exploiting its natural tendencies.
As training progresses, adjacent bins are sequentially introduced, expanding the range of movement directions the robot can reliably execute.
Performance tracking determines the progression of the curriculum.
Each bin is considered learned once the robot achieves at least 95\% of the target velocity over an episode.
Once the robot satisfies these conditions for a given bin, the curriculum introduces the next adjacent bin, continuing until the full circle of movement directions is covered.
When selecting a bin, a specific direction within it is uniformly sampled to ensure full coverage.

\subsection{Filter and reward ablations}
\label{app:ablations}
\begin{figure}[h]
    \centering
    \includegraphics[width=0.49\linewidth]{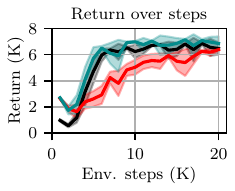}
    \hfill
    \includegraphics[width=0.49\linewidth]{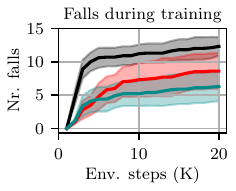}
    \includegraphics[width=\linewidth]{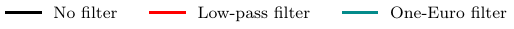}
    \caption{
        Ablation study of filter setups, illustrating their impact on the average return (left) and the number of falls (right).
    }
    \label{fig:ablation_filter}
    \vspace{-0.5em}
\end{figure}

We carry out ablation studies to investigate the impact of different filter setups and reward terms on the learning performance.
The filter setups are compared in Figure \ref{fig:ablation_filter}, showing the average return and the number of falls for the One-Euro filter, the low-pass filter, and no filtering.
The One-Euro filter achieves a balance between the high performance of no filtering and the low amount of falls of the low-pass filter.

\begin{figure}[h]
    \centering
    \includegraphics[width=0.49\linewidth]{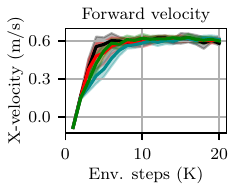}
    \hfill
    \includegraphics[width=0.49\linewidth]{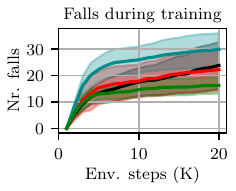}
    \includegraphics[width=\linewidth]{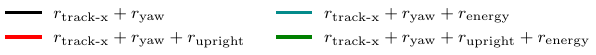}
    \caption{
        Ablation of reward terms, showing the impact on the achieved velocity and the number of falls during training.
    }
    \label{fig:ablation_reward}
    \vspace{-0.5em}
\end{figure}

Different compositions of reward terms are compared in Figure \ref{fig:ablation_reward}, illustrating the impact of additional reward penalties, especially on the number of falls during training, while maintaining the same target velocity.
The additional penalties significantly reduce the amount of falls, achieving a 50\% reduction compared to only using the tracking reward terms.

\clearpage
\section*{ACKNOWLEDGMENT}
We sincerely thank MAB Robotics for providing the HoneyBadger robot and all the extensive support and guidance during the experiments.
We especially thank Jakub Matyszczak, Jakub Bartoszek, Krzysztof Kwapisz and Michał Raszewski.


\bibliographystyle{IEEEtran}
\bibliography{IEEEabrv,bibliography}

\end{document}